\pdfoutput=1

\documentclass[11pt]{article}

\usepackage[final]{coling}

\usepackage{times}
\usepackage{latexsym}

\usepackage[T1]{fontenc}

\usepackage[utf8]{inputenc}

\usepackage{microtype}

\usepackage{inconsolata}

\usepackage{graphicx}
\usepackage{amsmath}
\usepackage{multirow}
\usepackage{array}
\usepackage[normalem]{ulem}
\useunder{\uline}{\ul}{}

\title{Attention-Seeker: Dynamic Self-Attention Scoring for Unsupervised Keyphrase Extraction}

\author{
 \textbf{Erwin D. López Z.\textsuperscript{1}},
 \textbf{Cheng Tang\textsuperscript{2}},
 \textbf{Atsushi Shimada\textsuperscript{2}},
\\
 Graduate School of Information Science and Electrical Engineering, Kyushu University,
\\
\textsuperscript{1}\texttt{edlopez96s6@gmail.com}
\\
\textsuperscript{2}\texttt{\{tang,atsushi\}@ait.kyushu-u.ac.jp}
}

\begin{document}
\maketitle
\begin{abstract}
This paper proposes Attention-Seeker, an unsupervised keyphrase extraction method that leverages self-attention maps from a Large Language Model to estimate the importance of candidate phrases. Our approach identifies specific components – such as layers, heads, and attention vectors – where the model pays significant attention to the key topics of the text. The attention weights provided by these components are then used to score the candidate phrases. Unlike previous models that require manual tuning of parameters (e.g., selection of heads, prompts, hyperparameters), Attention-Seeker dynamically adapts to the input text without any manual adjustments, enhancing its practical applicability. We evaluate Attention-Seeker on four publicly available datasets: Inspec, SemEval2010, SemEval2017, and Krapivin. Our results demonstrate that, even without parameter tuning, Attention-Seeker outperforms most baseline models, achieving state-of-the-art performance on three out of four datasets, particularly excelling in extracting keyphrases from long documents. The source code of Attention-Seeker is available at \url{https://github.com/EruM16/Attention-Seeker}.
\end{abstract}

\section{Introduction}

Keyphrase extraction is a critical task in Natural Language Processing (NLP) where the core content of documents is summarized into a set of words or phrases. This process facilitates efficient and accurate information retrieval, which is valuable for several NLP applications, including Retrieval-Augmented Generation (RAG), document categorization, text segmentation, and topic modeling. Keyphrase extraction methods are generally classified into two categories: supervised and unsupervised. Supervised methods achieve high performance by harnessing the power of deep learning techniques, such as neural networks \citep{mult_per},  LSTM \cite{sahrawa}, GNN \cite{divgraph}, and transformers \cite{tnt_kid}. However, these methods require large amounts of labeled data and are often domain-specific, limiting their practical applicability. In contrast, unsupervised methods rely solely on information extracted from the document itself, making them adaptable across various domains.

Unsupervised keyphrase extraction methods can be divided into five categories based on the type of information they extract. The first two categories include traditional approaches: statistical methods \citep{tf_idf_retr,yake}, which use in-corpus statistical information, and graph-based methods \citep{textrank,singlerank,topicrank,positionrank}, which leverage words co-occurrence patterns. The remaining three categories consist of more recent works that use Pretrained Language Models (PLMs) to extract semantic information. Embedding-based methods \cite{embedrank,sifrank,attentionrank,mderank,jointgl} analyze similarities between documents and phrases within the PLM's embedding space. Prompt-based methods \citep{promptrank} use the PLM's decoder logits to estimate the probability of generating a candidate phrase when a prompt is included in the document (e.g., this book talks about [candidate]). Finally, self-attention-based methods \citep{samrank} examine the PLMs' Self-Attention Map (SAM) to identify which candidate phrases receive the most attention across all document tokens.

As noted by \citet{promptrank,samrank}, embedding-based approaches struggle to accurately estimate document similarities due to the anisotropic nature of PLM embedding spaces. Prompt and self-attention-based methods circumvent these limitations and achieve state-of-the-art (SOTA) results but introduce new complexities. For example, PromptRank \citep{promptrank}, a prompt-based method, requires tuning two hyperparameters (alpha and beta) and exploring a wide range of potential prompts. Similarly, SAMRank \citep{samrank}, a self-attention-based model, requires selecting an optimal SAM from potentially thousands of layer and head combinations. Since documents typically lack labels to guide parameter selection, achieving optimal results without parameterization remains a significant challenge.

In this context, we propose a method that automatically adapts to the characteristics of the input text, eliminating the need for manual parameter tuning. Our approach extends SAMRank \citep{samrank} by introducing a module that selects the most relevant SAMs to effectively score candidate phrases. The relevance of SAMs is estimated by defining the characteristics of an optimal SAM. Specifically, Attention-Seeker employs a binary vector that assigns ones to candidate phrases and zeros to all other tokens. It then computes the similarity between this vector and the attention vectors (rows) of the SAMs, using the average similarity score to determine each SAM’s relevance. Furthermore, Attention-Seeker evaluates the relevance of different parts of the text: for short documents, it scores individual attention vectors, whereas for long documents, it scores document segments.

To the best of our knowledge, Attention-Seeker is the first method to automatically adapt its keyphrase extraction process to the specific characteristics of the input text. This novel approach allows Attention-Seeker to achieve SOTA-level performance on four benchmark datasets without requiring parameter tuning. Our ablation study shows that each of the Attention-Seeker's scoring modules positively contributes to the final performance. While our proposed relevance scoring mechanism is simple, its effectiveness highlights the potential for further exploration in this direction. 

The main contributions of this paper are summarized as follows:
\begin{itemize}
    \item We propose Attention-Seeker, a novel unsupervised keyphrase extraction method that automates parameter tuning of self-attention-based models.
    \item We propose a simple yet effective approach for selecting the most relevant attention vectors, self-attention maps, and document segments for keyphrase extraction. 
    \item We demonstrate that non-parametric self-attention-based methods using LLAMA 3-8B achieve high performance on four benchmark datasets, with notable efficacy on long documents.
\end{itemize}

\section{Related work}

\subsection{Unsupervised Keyphrase Extraction}
Traditional unsupervised keyphrase extraction methods can be broadly divided into statistic-based and graph-based approaches. Statistic-based methods estimate word importance using statistical metrics derived from the text. For instance, the TF-IDF method \citep{tf_idf_retr} relies on word frequencies, while YAKE \citep{yake} incorporates additional factors such as word co-occurrences, positions, and casings. In contrast, graph-based methods represent documents as graphs, with words as nodes, and use graph algorithms to estimate word importance. TextRank \citep{textrank} pioneered this approach by adapting the PageRank algorithm \citep{pagerank} for keyphrase extraction. Subsequent work has extended this framework: SingleRank \citep{singlerank} incorporates information from neighboring documents, TopicRank \citep{topicrank} clusters nodes within a topic space, and PositionRank \citep{positionrank} integrates a position bias into the PageRank algorithm.

Recent advances use Pretrained Language Models (PLMs) to extract semantic features from documents. These approaches can be categorized into embedding-based, prompt-based, and self-attention-based methods. Embedding-based approaches are the most common, focusing on generating embedding vectors from documents and candidate keyphrases. For example, EmbedRank \citep{embedrank} uses Doc2Vec \citep{doc2vec} and Sent2Vec \citep{sent2vec} to derive document and phrase embeddings, using cosine similarity for ranking. SIFRank \citep{sifrank} extends this approach by integrating ELMo \citep{elmo} embeddings and the SIF model \citep{sif} to refine document embeddings. JointGL \citep{jointgl} applies a similar strategy using BERT \citep{bert} embeddings while incorporating local similarities derived from document graphs. AttentionRank \citep{attentionrank} introduces cross-attention with BERT embeddings to generate document vectors, using self-attention maps (SAMs) for further refinement. Finally, MDERank \citep{mderank} takes an innovative approach by masking candidate tokens and comparing their similarity to the original document. 

Prompt-based methods take a different approach, using PLMs to predict keyphrases directly through prompt conditioning. For example, PromptRank \cite{promptrank} estimates the likelihood of candidate phrases by using the logits generated by the T5 \cite{t_5} PLM, which is specifically prompted for keyphrase extraction. In contrast, self-attention-based methods, such as SAMRank \citep{samrank}, focus on using SAMs from PLMs. SAMRank aggregates attention vectors from a specific SAM to measure the attention a candidate phrase receives from other tokens. While SAMRank demonstrates the usefulness of SAMs in keyphrase extraction, it highlights the challenge of effectively selecting the most relevant SAMs without label information.

\subsection{Self-Attention Map}
\citet{att_is_all} introduced the "Scaled Dot-Product Attention" mechanism, as shown in Equation ~\ref{eq:eq_0}. Here, the Query (Q), Key (K), and Value (V) are linear transformations of the embedding representation of the same input sentence. The Softmax function assigns an attention score to the tokens in the Key based on their relevance to a token in the Query, resulting in a matrix known as the Self-Attention Map (SAM). This SAM is used to compute a weighted sum of the Value vectors, which is then transformed into a new embedding representation of the input sentence for further processing by the same attention mechanism in subsequent layers.

\begin{equation}
\label{eq:eq_0}
    Att\left ( Q,K,V \right )=Softmax\left ( \frac{QK^{T}}{\sqrt{d_{k}}} \right )V
\end{equation}

 In the transformer model proposed by \citet{att_is_all}, and in subsequent PLMs, each layer contains multiple heads. As a result, each SAM can focus on different aspects of text processing (multiple heads) at different syntactic levels (multiple layers). \citet{bert_att} examined the characteristics of different SAMs in the BERT model \citep{bert} and identified several types: SAMs that specialize in attending to previous or next tokens, SAMs that focus on separator tokens (e.g., special tokens, periods, and commas), and SAMs that attend to a large number of words (uniform attention). Furthermore, they identified specific SAMs that preserve syntactic information, suggesting that some SAMs might specialize in attending to keyphrases within the text. Supporting this hypothesis, \citet{samrank} demonstrated that attention scores from a manually selected SAM can be leveraged to achieve SOTA performance in keyphrase extraction tasks. However, identifying and selecting SAMs specialized in attending to keyphrases remains an open challenge in this approach.
 
 \section{Methodology}
 
In this section, we introduce our proposed method, Attention-Seeker. It consists of four main steps: (1) Generation of candidate phrases using Part-of-Speech (POS) sequences. (2) Extraction of Self-Attention Maps (SAMs). (3) Estimation of a vector of attention scores from the most relevant SAMs. (4) Scoring the importance of each candidate phrase based on these attention scores. The main contribution of our method lies in step 3, where we propose a novel approach to estimate the vector of attention scores by identifying and exploiting information from the most relevant SAMs.

\subsection{Candidate Generation}
We follow a well-established approach for generating candidate phrases \citep{samrank,attentionrank,embedrank}. First, we tokenize and POS tag all words in the document by using the Stanford CoreNLP tool\footnote{\url{https://stanfordnlp.github.io/CoreNLP/}}. Next, we extract noun phrases (tagged as “NN”, “NNS”, “NNP”, “NNPS”, and “JJ”) using the NLTK’s RegexpParser\footnote{\url{https://github.com/nltk}}. These phrases are then defined as the document's keyphrase candidates.

\begin{figure*}[h]
  \centering
  \includegraphics[width=0.8\linewidth]{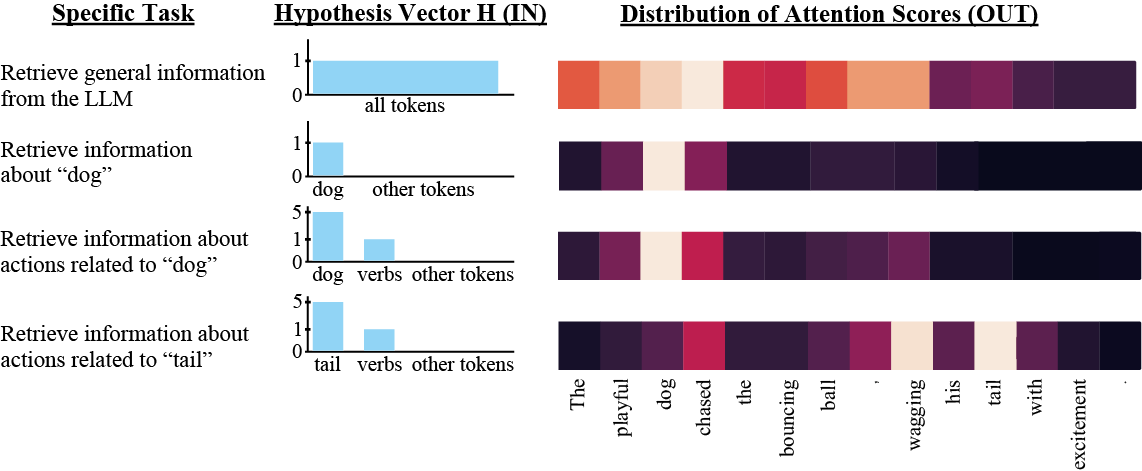}
  \caption{The effect of different hypothesis vectors H in an LLM's distribution of attention scores.}
  \label{fig:fig_0}
\end{figure*}

\subsection{Extraction of the Self-Attention Maps}
\label{sec:extraction_SAMs}
We extract the SAMs from each layer and head of the Large Language Model LLAMA 3-8B \citep{llama3} using the Huggingface library\footnote{\url{https://github.com/huggingface/transformers}}. Similar to SAMRank \citep{samrank}, we handle the extraction of SAMs differently depending on the document length. For short documents, we input the entire text directly into the LLM. For long documents, we first input the document's abstract into the LLM, then divide the remaining text into equally sized segments (same number of tokens) and process each segment independently with the LLM. 

\subsection{SAMs' Relevance Scoring: Hypothesis Engineering}
All Self-Attention Maps (SAMs) in a Large Language Model (LLM) capture distinct attention patterns over tokens, and their distribution of attention scores serves as a unique fingerprint. We propose to use these fingerprints to measure the relevance of each SAM within the LLM. Given a SAM of size $n  \times n$, our method requires the design of a hypothesis vector $H$ of length $n$ that encapsulates the desired properties of the attention distribution for a given task. Then, we calculate a similarity score between the SAM and the vector $H$ that quantifies the relevance of the SAM to the given task. We use this score as a weight to combine the attention distributions of all SAMs into a single output distribution for the LLM. This averaged distribution aggregates information across the different contexts captured by all layers and heads, prioritizing the contributions of the most relevant SAMs.

Figure~\ref{fig:fig_0} illustrates the effect of a hypothesis vector $H$ on the output distribution of attention scores for the same input text. The first case represents the default output distribution of the LLM, where all SAMs are considered equally relevant. In this scenario, $H$ is defined as a uniform vector with all elements set to $1$, resulting in equal similarity scores for all SAMs. In the second case, $H$ is defined as a binary vector where only the elements corresponding to “dog” are set to $1$. This configuration assigns greater relevance to SAMs that focus on syntactic relationships involving the noun “dog” and facilitates the retrieval of information about “dog”. A simple post-processing step, which masks the attention scores of “dog” from the final distribution, allows the extraction of associated words such as “playful” and “chased”.

The third and fourth cases demonstrate the use of combined hypothesis vectors. Both cases include a hypothesis vector designed to prioritize SAMs that capture syntactic information related to verbs (assigning $1$ to tokens corresponding to verbs and $0$ to others). In the third case, this verb-focused vector is combined with the hypothesis vector for “dog”, while in the fourth case, it is combined with the vector for “tail”. As shown in Figure~\ref{fig:fig_0}, the resulting distributions differ significantly, highlighting the verb most related to “dog” in the third case and the verb most related to “tail” in the fourth case. These examples illustrate how task-specific hypothesis vectors can guide the extraction of syntactically relevant information.

Although the examples shown in Figure~\ref{fig:fig_0} use simple hypothesis vectors, future works could explore advanced methodologies to design optimal vectors tailored to specific tasks. In this study, we propose two simple designs for keyphrase extraction: one for short and another for long documents.  Both designs start from a vector $H$ that focuses on the keyphrase candidates. This approach prioritizes SAMs that model the inter-relationships of these candidates. Among the relevant SAMs, we can include those capturing long-term dependencies, object-direct relationships, coreferences, and keyphrases. 

\begin{figure*}[h]
  \centering
  \includegraphics[width=\linewidth]{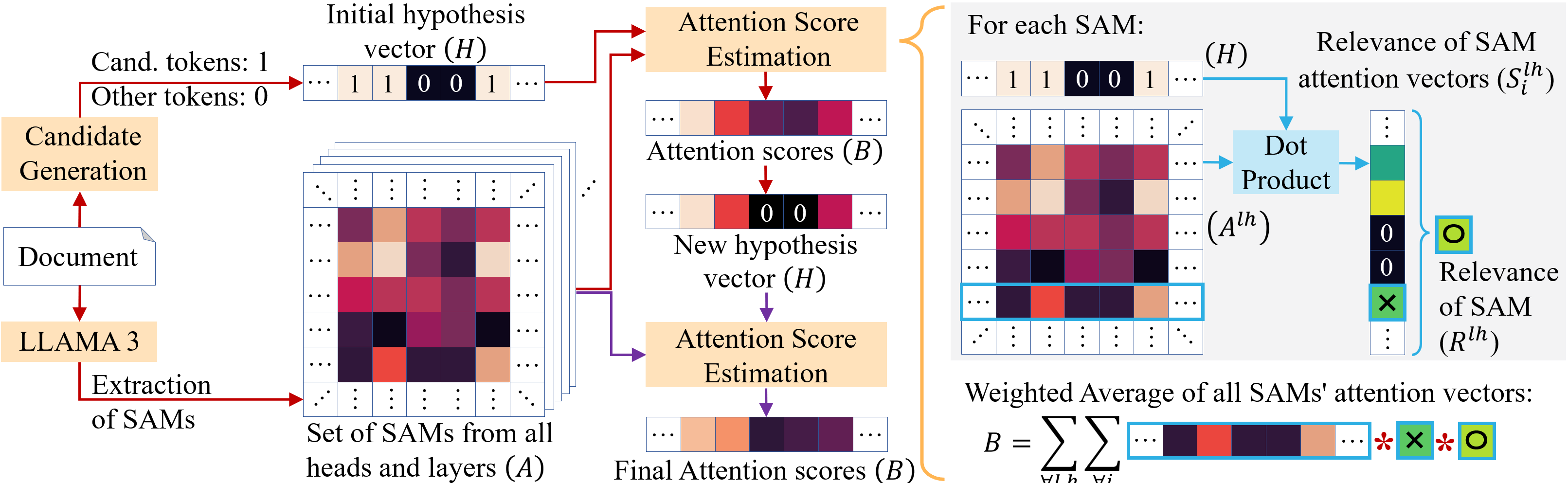}
  \caption{The core architecture of Attention-Seeker for short documents.}
  \label{fig:fig_1}
\end{figure*}

\subsection{Attention Scores Estimation: Short Documents}
\label{sec:short_att}
Our proposed method for short documents is summarized in Figure~\ref{fig:fig_1}. Here, the "Attention Score Estimation" module first estimates the relevance of SAMs ($R^{lh}$) and their internal attention vectors ($S^{lh}_{i}$). These relevance values are then used to compute a weighted average of the attention vectors across all SAMs. Accordingly, the final vector of attention scores $B$ is determined from the most relevant parts of the most relevant SAMs.

We define the hypothesis vector $H$ as in Equation~\ref{eq:eq_1}. This vector prioritizes SAMs that pay attention to potential keyphrases (candidates), while de-emphasizing those that primarily attend to token positions or separator tokens \citep{bert_att}.

\begin{equation}
  \label{eq:eq_1}
H_i =\begin{cases}
1, &\text{token } i \in \text{candidate}\\
0, &\text{otherwise}
\end{cases}
\end{equation}

Given a SAM $A^{lh}$, where $l$ is the current layer and $h$ is the current head, the relevance of each attention vector $S^{lh}_{i}$ is calculated as the dot product between $H$ and $A^{lh}_{i}$. This calculation is formally expressed as the matrix multiplication shown in Equation~\ref{eq:eq_2}, where $S^{lh}$ is the vector of $S^{lh}_{i}$ values.

\begin{equation}
  \label{eq:eq_2}
S^{lh}=A^{lh}\cdot H
\end{equation}

We further set $S^{lh}_{i}$ to $0$ for tokens $i$ that do not belong to candidate phrases (post-processing masking). This is an empirical decision, and can be justified by the existence of uniformly distributed attention vectors (e.g., periods preserving contextual information for subsequent sentences) or randomly distributed attention vectors for "non-op" contexts \citep{bert_att}.

As shown by Equation~\ref{eq:eq_3}, the relevance of each SAM $R^{lh}$ is calculated as the average relevance score of its individual attention vectors.

\begin{equation}
  \label{eq:eq_3}
 R^{lh} = \frac{1}{n}\sum_{i=0}^{n}S^{lh}_{i}
\end{equation}

The final attention vector $B$ is computed as a weighted average of all attention vectors across all SAMs, with the weights determined by their respective relevance scores, as shown in Equation~\ref{eq:eq_4}.

\begin{equation}
  \label{eq:eq_4}
B= \sum_{\forall l,h} \sum_{\forall i} A^{lh}_{i}*S^{lh}_{i}*R^{lh}
\end{equation}

As illustrated in Figure~\ref{fig:fig_1}, the resulting vector $B$ is converted into a new hypothesis vector $H$ through post-processing masking, where $B_{i}$ values corresponding to non-candidate tokens are set to 0. Similar to the initial binary vector $H$, this updated vector prioritizes SAMs that focus on possible keyphrases. However, it further emphasizes SAMs that allocate more attention to the most important candidates, thereby favoring SAMs likely to specialize in attending to keyphrases.

\subsection{Attention Scores Estimation: Long Documents}

\begin{figure*}[h]
  \centering
  \includegraphics[width=\linewidth]{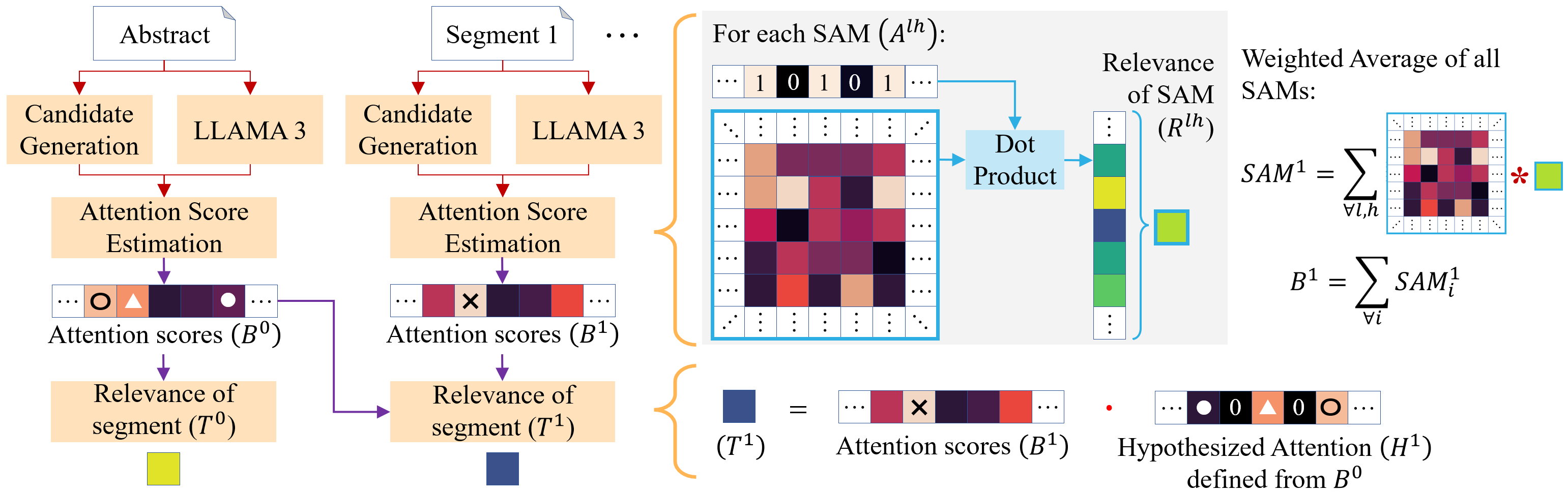}
  \caption{The core architecture of Attention-Seeker for long documents.}
  \label{fig:fig_2}
\end{figure*}

Long documents are segmented into an abstract and equally sized segments, as described in Section~\ref{sec:extraction_SAMs}. These segments are processed independently as short documents (Section~\ref{sec:short_att}) excluding the relevance scores for attention vectors ($S^{lh}_{i}$) from the final calculation (Figure~\ref{fig:fig_2}). This simplification avoids the complexities of applying context-dependent relevance scores across diverse segments, which could reduce effectiveness without a robust normalization strategy.

For each segment $s$, the relevance of each SAM ($R^{lh}$) is defined as the average of the relevance scores of its corresponding attention vectors (Equation ~\ref{eq:eq_5}). 

\begin{equation}
  \label{eq:eq_5}
 R^{lh} = \frac{1}{n}\sum_{i=0}^{n}A^{lh}_{i}\cdot H
\end{equation}

We then apply $l_{1}$-normalization to this vector (Equation~\ref{eq:eq_6}) to compute a weighted average $SAM^{s}$ of all SAMs from the segment $s$ (Equation~\ref{eq:eq_7}). This normalization ensures that $SAM^{s}$ preserves the standard properties of a SAM (e.g., the attention values in each row sum to 1). 

\begin{equation}
  \label{eq:eq_6}
 R^{lh}_{n} = \frac{R^{lh}}{\sum_{\forall l,h}R^{lh}}
\end{equation}

\begin{equation}
  \label{eq:eq_7}
SAM^{s}= \sum_{\forall l,h} A^{lh}*R^{lh}_{n}
\end{equation}

The attention score vector $B^{s}$ for the segment $s$ is obtained by summing all rows in the average SAM (Equation~\ref{eq:eq_8}).

\begin{equation}
  \label{eq:eq_8}
B^{s}= \sum_{\forall i} SAM^{s}_{i}
\end{equation}

To compute the relevance of each segment ($T^{s}$), we define a new hypothesis vector $H^{s}$ for each segment. This vector assigns the attention scores from the abstract ($B^{0}$) to tokens in the segment that are included in the candidate phrases of the abstract, while assigning 0 to all other tokens (Equation ~\ref{eq:eq_9}). This hypothesis vector prioritizes segments that focus more on phrases considered important by the abstract.

\begin{equation}
  \label{eq:eq_9}
H_i^{s} =\begin{cases}
B^{0}_{j}, &\text{token } i = \text{token } j \text{ of  }B^{0}\\
0, &\text{otherwise}
\end{cases}
\end{equation}

The relevance score $T^{s}$ for each segment $s$ is finally calculated as the dot product between its attention score vector $B^{s}$ and the hypothesis vector $H^{s}$ (Equation ~\ref{eq:eq_10}).

\begin{equation}
  \label{eq:eq_10}
T^{s}=B^{s}\cdot H^{s}
\end{equation}

\subsection{Candidate Final Score Calculation}
In short documents, the final score of a candidate phrase is calculated by summing the attention scores $B_{i}$ corresponding to the candidate tokens. Following \citet{samrank}, if a candidate phrase consists of a single word, its final score is divided by its frequency. For long documents, we apply the same method to each segment with the vector of attention scores $T^{s}*B^{s}$ ($B^{s}$ multiplied by the relevance of the segment). The final score for each candidate is then obtained by summing its scores across all segments.

\begin{table}[b]
\fontsize{10pt}{13pt}\selectfont
\addtolength{\tabcolsep}{-2pt}
\centering
\begin{tabular}{l|cccc}
\hline
             & Inspec & SE2017 & SE2010 & Krapivin \\ \hline
\# Docs      & 500    & 493    & 100    & 460      \\
Ave. Words   & 135    & 194    & 8154   & 8545     \\
Ave. Sent.   & 6      & 7      & 380    & 312      \\
Ave. Keys    & 9      & 17     & 15     & 6        \\
Unigram (\%) & 13.5   & 25.7   & 20.5   & 17.8     \\
Bigram (\%)  & 52.7   & 34.4   & 53.6   & 62.2     \\
Trigram (\%) & 24.9   & 17.5   & 18.9   & 16.4     \\ \hline
\end{tabular}
\caption{Statistics of datasets}
\label{tab:tab_1}
\end{table}

\begin{table*}[h]
\fontsize{10pt}{13pt}\selectfont
\setlength{\ULdepth}{0.25ex} 
\addtolength{\tabcolsep}{-4pt}
\centering
\begin{tabular}{lcccccccccccc}
\hline
\multicolumn{1}{c|}{\multirow{2}{*}{\textbf{Method}}} &
  \multicolumn{3}{c|}{\textbf{Inspec}} &
  \multicolumn{3}{c|}{\textbf{SemEval2017}} &
  \multicolumn{3}{c|}{\textbf{SemEval2010}} &
  \multicolumn{3}{c}{\textbf{Krapivin}} \\ \cline{2-13} 
\multicolumn{1}{c|}{} &
  F1@5 &
  F1@10 &
  \multicolumn{1}{c|}{F1@15} &
  F1@5 &
  F1@10 &
  \multicolumn{1}{c|}{F1@15} &
  F1@5 &
  F1@10 &
  \multicolumn{1}{c|}{F1@15} &
  F1@5 &
  F1@10 &
  F1@15 \\ \hline
\multicolumn{13}{c}{Statistic-based Methods} \\ \hline
\multicolumn{1}{l|}{TF-IDF} &
  11.28 &
  13.88 &
  \multicolumn{1}{c|}{13.83} &
  12.70 &
  16.26 &
  \multicolumn{1}{c|}{16.73} &
  2.81 &
  3.48 &
  \multicolumn{1}{c|}{3.91} &
  - &
  - &
  - \\
\multicolumn{1}{l|}{YAKE} &
  18.08 &
  19.62 &
  \multicolumn{1}{c|}{20.11} &
  11.84 &
  18.14 &
  \multicolumn{1}{c|}{20.55} &
  11.76 &
  14.40 &
  \multicolumn{1}{c|}{15.19} &
  8.09 &
  9.35 &
  11.05 \\ \hline
\multicolumn{13}{c}{Graph-based Methods} \\ \hline
\multicolumn{1}{l|}{TextRank} &
  27.04 &
  25.08 &
  \multicolumn{1}{c|}{36.65} &
  16.43 &
  25.83 &
  \multicolumn{1}{c|}{30.50} &
  3.80 &
  5.38 &
  \multicolumn{1}{c|}{7.65} &
  6.04 &
  9.43 &
  9.95 \\
\multicolumn{1}{l|}{SingleRank} &
  27.79 &
  34.46 &
  \multicolumn{1}{c|}{36.05} &
  18.23 &
  27.73 &
  \multicolumn{1}{c|}{31.73} &
  5.90 &
  9.02 &
  \multicolumn{1}{c|}{10.58} &
  8.12 &
  10.53 &
  10.42 \\
\multicolumn{1}{l|}{TopicRank} &
  25.38 &
  28.46 &
  \multicolumn{1}{c|}{29.49} &
  17.10 &
  22.62 &
  \multicolumn{1}{c|}{24.87} &
  12.12 &
  12.90 &
  \multicolumn{1}{c|}{13.54} &
  8.94 &
  9.01 &
  8.30 \\
\multicolumn{1}{l|}{PositionRank} &
  28.12 &
  32.87 &
  \multicolumn{1}{c|}{33.32} &
  18.23 &
  26.30 &
  \multicolumn{1}{c|}{30.55} &
  9.84 &
  13.34 &
  \multicolumn{1}{c|}{14.33} &
  - &
  - &
  - \\ \hline
\multicolumn{13}{c}{Embedding-based Methods} \\ \hline
\multicolumn{1}{l|}{EmbedRank d2v} &
  31.51 &
  37.94 &
  \multicolumn{1}{c|}{37.96} &
  20.21 &
  29.59 &
  \multicolumn{1}{c|}{33.94} &
  3.02 &
  5.08 &
  \multicolumn{1}{c|}{7.23} &
  4.05 &
  6.60 &
  7.84 \\
\multicolumn{1}{l|}{EmbedRank s2v} &
  29.88 &
  37.09 &
  \multicolumn{1}{c|}{38.40} &
  - &
  - &
  \multicolumn{1}{c|}{-} &
  5.40 &
  8.91 &
  \multicolumn{1}{c|}{10.06} &
  8.44 &
  10.47 &
  10.71 \\
\multicolumn{1}{l|}{SIFRank} &
  29.11 &
  {\ul 38.80} &
  \multicolumn{1}{c|}{{\ul \textbf{39.59}}} &
  22.59 &
  32.85 &
  \multicolumn{1}{c|}{38.10} &
  - &
  - &
  \multicolumn{1}{c|}{-} &
  1.62 &
  2.52 &
  3.00 \\
\multicolumn{1}{l|}{AttentionRank} &
  24.45 &
  32.15 &
  \multicolumn{1}{c|}{34.49} &
  23.59 &
  34.37 &
  \multicolumn{1}{c|}{38.21} &
  11.39 &
  15.12 &
  \multicolumn{1}{c|}{16.66} &
  - &
  - &
  - \\
\multicolumn{1}{l|}{MDERank} &
  27.85 &
  34.36 &
  \multicolumn{1}{c|}{36.40} &
  20.37 &
  31.21 &
  \multicolumn{1}{c|}{36.63} &
  13.05 &
  18.27 &
  \multicolumn{1}{c|}{20.35} &
  11.78 &
  12.93 &
  12.58 \\
\multicolumn{1}{l|}{JointGL} &
  30.82 &
  36.28 &
  \multicolumn{1}{c|}{36.67} &
  20.49 &
  29.63 &
  \multicolumn{1}{c|}{34.05} &
  10.78 &
  13.67 &
  \multicolumn{1}{c|}{14.64} &
  - &
  - &
  - \\ \hline
\multicolumn{13}{c}{Prompt-based Methods} \\ \hline
\multicolumn{1}{l|}{PromptRank} &
  31.73 &
  37.88 &
  \multicolumn{1}{c|}{38.17} &
  {\ul \textbf{27.14}} &
  {\ul \textbf{37.76}} &
  \multicolumn{1}{c|}{{\ul \textbf{41.57}}} &
  17.24 &
  20.66 &
  \multicolumn{1}{c|}{21.35} &
  16.11 &
  16.71 &
  {\ul 16.02} \\ \hline
\multicolumn{13}{c}{Self-Attention Map-based Methods} \\ \hline
\multicolumn{1}{l|}{SAMRank} &
  {\ul 34.25} &
  38.18 &
  \multicolumn{1}{c|}{38.11} &
  24.74 &
  33.51 &
  \multicolumn{1}{c|}{37.01} &
  {\ul 17.86} &
  {\ul 20.99} &
  \multicolumn{1}{c|}{{\ul 22.07}} &
  {\ul 17.38} &
  {\ul 16.78} &
  15.15 \\
\multicolumn{1}{l|}{Attention-Seeker} &
  {\ul \textbf{35.49}} &
  {\ul \textbf{40.14}} &
  \multicolumn{1}{c|}{{\ul 39.22}} &
  {\ul 25.40} &
  {\ul 34.53} &
  \multicolumn{1}{c|}{{\ul 38.50}} &
  {\ul \textbf{19.00}} &
  {\ul \textbf{23.07}} &
  \multicolumn{1}{c|}{{\ul \textbf{23.81}}} &
  {\ul \textbf{20.79}} &
  {\ul \textbf{18.25}} &
  {\ul \textbf{16.22}} \\ \hline
\end{tabular}
\caption{
    Performance of keyphrase extraction of baseline models and Attention-Seeker on four datasets. The \textbf{bold} indicates the best performance and the \uline{underline} indicates the two best performances.
  }
\label{tab:tab_2}
\end{table*}

\section{Experiments and Results}
\subsection{Datasets and Evaluation Metrics}
We evaluate Attention-Seeker using four benchmark datasets: Inspec \citep{inspec}, SemEval2010 \citep{semeval2010}, SemEval2017 \cite{semeval2017}, and Krapivin \citep{Krapivin}. These datasets consist of scientific papers and vary in length. Specifically, Inspec and SemEval2017 contain paper abstracts (short documents), while SemEval2010 and Krapivin contain full papers (long documents). Table ~\ref{tab:tab_1} presents more detailed statistical information about each dataset.

The performance of keyphrase extraction is evaluated with the F1 score at the top 5 (F1@5), 10 (F1@10), and 15 (F1@15) keyphrases predicted by the model (candidates with the highest final scores). The final set of predicted keyphrases is determined after applying the NLTK's PorterStemmer for stemming and removing duplicate candidates.

\subsection{Baselines}
Our baselines include statistics-based methods: TF-IDF \citep{tf_idf_retr}, YAKE \citep{yake}; graph-based methods: TextRank \citep{textrank}, SingleRank \citep{singlerank}, TopicRank \citep{topicrank}, PositionRank \citep{positionrank}; embedding-based methods: EmbedRank \citep{embedrank}, SIFRank \citep{sifrank}, AttentionRank \citep{attentionrank}, MDERank \citep{mderank}, JointGL \citep{jointgl}; prompt-based methods: PromptRank \citep{promptrank}, and self-attention-based models: SAMRank \citep{samrank}. To ensure a fair comparison, we implemented SAMRank using the average of all SAMs (a non-parametric method) extracted with the LLAMA 3-8B model (BERT and GPT-2 results can be found on Appendix N). The optimized results of SAMRank are detailed in Appendix ~\ref{sec:apx_optimal_sam}. Additionally, we omitted the proportional score, as it led to lower performance in this setting (Appendix ~\ref{sec:apx_nonp_sam}). Consequently, the reported results of SAMRank represent the most important baseline for our model, as they correspond to the same method when all relevance scores are assigned equal values.

\begin{table*}[h]
\fontsize{10pt}{13pt}\selectfont
\addtolength{\tabcolsep}{-4.25pt}
\centering
\begin{tabular}{l|ccc|ccc|ccc|ccc}
\hline
\multicolumn{1}{c|}{\multirow{2}{*}{\textbf{Method}}} &
  \multicolumn{3}{c|}{\textbf{Inspec}} &
  \multicolumn{3}{c|}{\textbf{SemEval2017}} &
  \multicolumn{3}{c|}{\textbf{SemEval2010}} &
  \multicolumn{3}{c}{\textbf{Krapivin}} \\ \cline{2-13} 
\multicolumn{1}{c|}{}   & F1@5  & F1@10 & F1@15 & F1@5  & F1@10 & F1@15 & F1@5  & F1@10 & F1@15 & F1@5  & F1@10 & F1@15 \\ \hline
Base                    & 34.25 & 38.18 & 38.11 & 24.74 & 33.51 & 37.01 & 16.80 & 20.26 & 20.94 & 16.38 & 16.44 & 14.78 \\
$B+S^{lh}$              & 34.98 & 39.30 & 38.99 & 25.02 & 34.38 & 38.14 & 15.77 & 20.37 & 21.43 & 16.70 & 16.55 & 15.07 \\
$B+R^{lh}$              & 34.44 & 38.54 & 38.24 & 24.85 & 33.79 & 37.11 & 16.79 & 20.60 & 21.26 & 16.99 & 16.48 & 14.91 \\
$B+f$                   & 34.80 & 39.54 & 39.08 & 25.00 & 34.30 & 37.63 & 15.90 & 19.80 & 21.00 & 16.59 & 16.67 & 15.24 \\
$B+S^{lh}+R^{lh}$     & 35.07 & 39.56 & 39.05 & 25.25 & 34.57 & 38.17 & 16.63 & 20.41 & 21.36 & 16.92 & 16.58 & 15.25 \\
$B+S^{lh}+R^{lh}+f$ & 35.23 & 40.22 & 39.48 & 25.27 & 34.78 & 38.33 & 15.83 & 20.40 & 21.52 & 16.64 & 16.72 & 15.45 \\
Attention-Seeker                   & 35.49 & 40.14 & 39.22 & 25.40 & 34.53 & 38.50 & 16.74 & 20.26 & 21.73 & 17.14 & 16.89 & 15.27 \\ \hline
\end{tabular}
\caption{Performance of Attention-Seeker using different relevance scores in \textbf{short} documents.}
\label{tab:tab_3}
\end{table*}

\subsection{Results}
Table ~\ref{tab:tab_2} summarizes the performance of Attention-Seeker compared to baseline models across four benchmark datasets. The results show that Attention-Seeker achieves state-of-the-art (SOTA) performance on the Inspec, SemEval2010, and Krapivin datasets, while securing the second-best performance on the SemEval2017 dataset. The non-parametric SAMRank also performs well on all datasets, highlighting the effectiveness of Self-Attention Maps (SAMs) for unsupervised keyphrase extraction. In particular, the performance of self-attention-based methods on long document datasets (SemEval2017 and Krapivin) suggests a promising direction for extracting keyphrases from long documents using SAMs from LLMs. In this context, the consistent outperformance of Attention-Seeker over SAMRank demonstrates the advantages of considering the relevance of different SAMs and document segments when estimating the attention scores. 

Although PromptRank remains the best-performing method on the SemEval2017 dataset, it requires hyperparameter tuning for each specific benchmark \citep{promptrank}, which limits its suitability for applications without labeled data. Furthermore, \citet{promptrank,prompt_pre} demonstrate that the choice of prompts significantly affects its performance, requiring an extensive search to identify the optimal prompt. In contrast, Attention-Seeker achieves the second-best performance on this dataset without requiring parameter tuning. As \citet{samrank} note, different input texts activate different SAMs for keyphrase identification. Attention-Seeker adapts to these variations by analyzing the SAMs triggered by the input, positioning it as a robust and adaptable solution for unsupervised keyphrase extraction.

\section{Ablation study}
\subsection{Short documents}
Attention-Seeker integrates several modules designed to improve its overall performance. Specifically, it employs a relevance score for each SAM's attention vector ($S^{lh}$), a relevance score for each SAM ($R^{lh}$), and a filtering step $f$ applied to non-candidate tokens of $S^{lh}$ (Section ~\ref{sec:short_att}). To evaluate their contributions, we applied these modules separately. We also tested configurations using both scores $S^{lh}$ and $R^{lh}$, with and without the filtering step. Table ~\ref{tab:tab_3} presents the performance of Attention-Seeker in all these cases. In the "Base" configuration, all attention vectors and SAMs are assigned equal relevance, and no filtering is applied. The following configurations progressively incorporate the relevance scores ($S^{lh}$ and $R^{lh}$) and the filtering step ($f$) into the base method ($B$). The final version of Attention-Seeker corresponds to "$B+S^{lh}+R^{lh}+f$" applied twice (Figure ~\ref{fig:fig_1}). Performance on the SemEval2010 and Krapivin datasets is evaluated using only the abstracts of the papers (short documents).

Table ~\ref{tab:tab_3} shows that each module contributes positively to the overall performance of our method, with the $S^{lh}$ score providing the most significant improvements. While our proposed relevance scores are supported by the observations of \citet{bert_att}, the benefit of filtering attention vectors from certain tokens is less straightforward. \citet{bert_att} observed that in SAMs specialized in attending to verbs of direct objects, attention vectors of non-nouns often attend to the [SEP] token. They suggest that in such cases, this token might function as a "non-op", implying that the remaining attention scores would be randomly distributed (the model ignores this information). Our results in the "$B + f$" configuration suggest this hypothesis is correct, as the filtering step would be removing these noisy scores. However, results from SemEval2010 indicate that our filtering approach may also cause slight performance degradation, likely due to the simplicity of the current filtering mechanism. Therefore, future research should investigate how SAM's attention vectors differ based on different query tokens and determine which tokens produce more effective attention distributions for representing contextual information.

\begin{table*}[h]
\fontsize{10pt}{13pt}\selectfont
\centering
\begin{tabular}{l|ccc|ccc}
\hline
\multicolumn{1}{c|}{\multirow{2}{*}{\textbf{Method}}} & \multicolumn{3}{c|}{\textbf{SemEval2010}} & \multicolumn{3}{c}{\textbf{Krapivin}} \\ \cline{2-7} 
\multicolumn{1}{c|}{} & F1@5  & F1@10 & F1@15 & F1@5  & F1@10 & F1@15 \\ \hline
Base                  & 17.86 & 20.99 & 22.07 & 17.38 & 16.78 & 15.15 \\
$B_{as}$              & 17.35 & 21.04 & 21.05 & 17.05 & 16.50 & 14.83 \\
$B_{as}+R^{lh}$     & 17.34 & 21.27 & 21.83 & 17.44 & 16.79 & 15.17 \\
$B_{as}+T^{s}_{b}$  & 18.20 & 21.52 & 22.72 & 18.58 & 17.39 & 15.88 \\
$B_{as}+T^{s}$      & 19.05 & 22.75 & 23.69 & 20.49 & 17.94 & 16.01 \\
Attention-Seeker      & 19.00 & 23.07 & 23.81 & 20.79 & 18.25 & 16.22 \\ \hline
\end{tabular}
\caption{Performance of Attention-Seeker using different relevance scores in \textbf{long} documents.}
\label{tab:tab_4}
\end{table*}

\subsection{Long documents}
Attention-Seeker segments the document into abstract and equal-length segments of the remaining content. Since this approach slightly differs from the base method (\citealp{samrank}; the entire document is segmented into equal parts), we evaluated its impact on the final performance. In addition, we evaluated the contributions of the relevance score for each SAM ($R^{lh}$) and the relevance score for each document segment $T^{s}$. We considered two versions of this second relevance score: One uses our proposed refined vector $H^{s}$ ($T^{s}$), and the other uses a simple binary vector $H^{0}_{b}$ that assigns one to abstract candidates and zero otherwise ($T^{s}_{b}$). Table ~\ref{tab:tab_4} summarizes the results. In the "Base" and "$B_{as}$" configurations, all SAMs and segments are treated as equally relevant ("$B_{as}$" considers the abstract as the first segment). The "$B_{as}$" configuration is then extended with relevance scores ($R^{lh}$, $T^{s}_{b}$, and $T^{s}$), and the final Attention-Seeker corresponds to $B_{as} + R^{lh} + T^{s}$.

Table ~\ref{tab:tab_4} shows that our two proposed relevance scores contribute positively to the performance of Attention-Seeker, with the relevance of segments providing the most significant improvements. In addition, refining the hypothesis vector $H^{s}$ further improves the performance of the segment scores ($T^{s}>T^{s}_{b}$). Despite these advances, our segmentation approach introduces some performance degradation compared to the base model, likely due to the difference in length between the abstract and other segments. While the overall improvement achieved by our method outweighs the initial degradation, further research may consider exploring new methods for estimating segment relevance without compromising the performance of the base model.

Finally, the contribution of SAMs' relevance scores $R^{lh}$ in Table ~\ref{tab:tab_4} is consistent with the results in Table ~\ref{tab:tab_3}, suggesting that our approach for short documents is equally effective when applied to segments of long documents. This result suggests that the inclusion of the relevance score $S^{lh}$ has the potential to further improve the performance of Attention-Seeker for long documents. Further research should explore methods for normalizing attention scores across contexts that vary in syntax, number of nouns, topics, and other factors.

\subsection{Relevance of SAMs}

Attention-Seeker's score $R^{lh}$ identifies the SAMs most relevant for keyphrase extraction, eliminating the need for labeled data to select an optimal SAM. Our analysis of the $R^{lh}$ scores across samples from four datasets (Inspec, SemEval2017, SemEval2010, and Krapivin) reveals that SAMs from intermediate layers (9 to 15 out of 32), along with a few SAMs from the first and last layer contribute significantly to the keyphrase extraction task (see Appendix ~\ref{sec:apx_rel_sam}).

\section{Conclusion}
We proposed Attention-Seeker, a self-attention-based method for unsupervised keyphrase extraction that eliminates the need for manual parameter tuning. Our method assumes that certain Self-Attention Maps (SAMs) are specialized to focus on the most important phrases within a document. Accordingly, Attention-Seeker reframes keyphrase extraction as the task of identifying where the most crucial information is encapsulated in SAMs. For long documents, we extended this method by first identifying the most relevant segments from which to extract attention scores.

Attention-Seeker outperformed most baselines on four benchmark datasets, demonstrating the effectiveness of our approach. Since Attention-Seeker is the first method to estimate the relevance from the internal scores of a Large Language Model (LLM), its scoring mechanism is very simple and has potential for improvement, as suggested in our study. Future research should explore more sophisticated methods for relevance estimation, which could lead to further performance improvements. In addition, optimizing these relevance estimates may provide deeper insights into the internal processes of LLMs, potentially contributing to the design of more efficient pretrained language models and their application in unsupervised tasks.

\section{Limitations}
Attention-Seeker estimates the relevance of attention vectors by defining a set of desired characteristics and measuring how closely each attention vector matches them. However, the current implementation of this approach presents two major limitations. First, it uses the dot product between the desired vector and SAMs' attention vectors. While this method is effective, it could be improved by applying normalization techniques such as $l_{2}$-normalization (cosine similarity), softmax normalization (cross-attention), or softmax with temperature scaling. 

Second, the hypothesis vectors ($H$), which serve as reference vectors, are initially defined as binary vectors that focus on candidate phrases. This simplistic formulation may introduce noise into the attention distribution of the candidates and partially overlook important contextual information. Although we refine this vector ($H$) by using the attention scores extracted from the document ($B$), the initial estimation of $B$ still relies on the original binary vector. To address these limitations, future work could explore more sophisticated hypothesis vectors and improved methods for measuring their alignment with SAM's attention vectors.

Another important limitation is that the long document version of Attention-Seeker requires the first segment to be the abstract of the document. This condition limits the application of our proposed method to documents with an abstract. In addition, our ablation study showed that this approach negatively affects the overall performance of our method. Accordingly, we believe that further research should explore alternative methods for defining the hypothesis vector $H^{s}$ without relying on information extracted from the abstract. Possible solutions could include extracting the most important tokens from all segments under the assumption of equal relevance, and defining the hypothesis vector $H^{s}$ based on these tokens.

Finally, the implementation of our method is currently limited to open LLMs. Since Attention-seeker relies on the information provided by the Self-Attention Maps of Language Models, it cannot be applied to private models with restricted access to their weights, such as Open AI's GPT-4o/Open-o1 and Antropic's Claude models. This limitation is common in the field, as previous methods also rely on internal model information (e.g., SAMs, embeddings, logits). However, as open models continue to improve and close the performance gap with closed models, we believe our approach will provide valuable insights for future research.

\section*{Acknowledgments}
This work was supported by JST CREST Grant Number JPMJCR22D1 and JSPS KAKENHI Grant Number JP22H00551, Japan.


\bibliography{custom}

\appendix

\section{Optimal Performance of SAMRank on LLAMA 3-8B}
\label{sec:apx_optimal_sam}

We adapted the original SAMRank method \citep{samrank} from the official repository \url{https://github.com/kangnlp/SAMRank}, using the LLAMA 3-8B model instead of GPT-2. Due to the structural similarities between GPT-2 and LLAMA 3-8B, no major modifications were required. Table ~\ref{tab:tab_apx1} shows the results obtained by the optimal SAMs on four benchmark datasets. We evaluated the performance of the method using both Global ($G$) and Final ($F$) scores.

\begin{table}[h]
\fontsize{10pt}{13pt}\selectfont
\addtolength{\tabcolsep}{-3pt}
\centering
\begin{tabular}{lcccc}
\hline
\multicolumn{1}{l|}{\textbf{Dataset}} & \textbf{F1@5} & \textbf{F1@10} & \multicolumn{1}{c|}{\textbf{F1@15}} & \textbf{SAM} \\ \hline
\multicolumn{5}{c}{SAMRank Final Score (S)}                                                        \\ \hline
\multicolumn{1}{l|}{\textbf{Inspec}}      & 33.19 & 39.04 & \multicolumn{1}{c|}{39.50} & L:19 H:15 \\
\multicolumn{1}{l|}{\textbf{SemEval2017}} & 24.50 & 34.05 & \multicolumn{1}{c|}{37.65} & L:21 H:11 \\
\multicolumn{1}{l|}{\textbf{SemEval2010}} & 16.30  & 18.04  & \multicolumn{1}{c|}{19.74}  & L:12 H:2 \\
\multicolumn{1}{l|}{\textbf{Krapivin}}    & 18.55  & 17.25  & \multicolumn{1}{c|}{15.63}  & L:8 H:8  \\ \hline
\multicolumn{5}{c}{SAMRank Global Score (G)}                                                       \\ \hline
\multicolumn{1}{l|}{\textbf{Inspec}}      & 35.60 & 40.47 & \multicolumn{1}{c|}{40.21} & L:9 H:7   \\
\multicolumn{1}{l|}{\textbf{SemEval2017}} & 25.64 & 36.41 & \multicolumn{1}{c|}{39.80} & L:20 H:28 \\
\multicolumn{1}{l|}{\textbf{SemEval2010}} & 17.10  & 19.32  & \multicolumn{1}{c|}{20.40}  & L:12 H:2 \\
\multicolumn{1}{l|}{\textbf{Krapivin}}    & 19.18  & 17.60  & \multicolumn{1}{c|}{15.93}  & L:12 H:2  \\ \hline
\end{tabular}
\caption{
    Performance of keyphrase extraction of SAMRank (Global and Final scores) on four datasets and their corresponding heads.
  }
\label{tab:tab_apx1}
\end{table}

The performance of the final score in SAMRank LLAMA 3-8B is consistent with the results reported by \citet{samrank} for SAMRank using GPT-2, LLAMA 2-7B and LLAMA 2-13B. However, in our implementation, the independent Global Score provides the best results, suggesting that the Proportional Score may be less effective in this context. While the ablation study conducted by \citet{samrank} demonstrated the benefits of the Proportional Score for BERT and GPT-2 implementations, its effectiveness for larger LLMs such as LLAMA remains unclear. We believe that further research is needed to determine the impact of the Proportional Score on LLM implementations of SAMRank.

The independent implementation of the Global Score in SAMRank outperforms Attention-Seeker in short documents, suggesting room for improvement in our method. As discussed in the main paper, these improvements could include more effective strategies for defining the desired characteristics of attention vectors and better measuring their alignment with these characteristics. In future implementations, we might expect results similar to those reported in Table ~\ref{tab:tab_apx1} (Global Score). However, it is important to note that these results may be inflated by non-relevant SAMs achieving high performance by chance, due to suboptimal labels and the large number of 1024 possible trials. 

\begin{table*}[h]
\fontsize{10pt}{13pt}\selectfont
\setlength{\ULdepth}{0.25ex} 
\addtolength{\tabcolsep}{-4pt}
\centering
\begin{tabular}{lcccccccccccc}
\hline
\multicolumn{1}{c|}{\multirow{2}{*}{\textbf{Method}}} &
  \multicolumn{3}{c|}{\textbf{Inspec}} &
  \multicolumn{3}{c|}{\textbf{SemEval2017}} &
  \multicolumn{3}{c|}{\textbf{SemEval2010}} &
  \multicolumn{3}{c}{\textbf{Krapivin}} \\ \cline{2-13} 
\multicolumn{1}{c|}{} &
  F1@5 &
  F1@10 &
  \multicolumn{1}{c|}{F1@15} &
  F1@5 &
  F1@10 &
  \multicolumn{1}{c|}{F1@15} &
  F1@5 &
  F1@10 &
  \multicolumn{1}{c|}{F1@15} &
  F1@5 &
  F1@10 &
  F1@15 \\ \hline
\multicolumn{13}{c}{BERT Model} \\ \hline
\multicolumn{1}{l|}{SAMRank (B)} &
  32.43 &
  38.26 &
  \multicolumn{1}{c|}{38.84} &
  22.93 &
  32.14 &
  \multicolumn{1}{c|}{35.85} &
  15.19 &
  18.70 &
  \multicolumn{1}{c|}{20.21} &
  11.99 &
  12.51 &
  12.14 \\
\multicolumn{1}{l|}{Attention-Seeker} &
  \textbf{34.54} &
  \textbf{40.31} &
  \multicolumn{1}{c|}{\textbf{40.15}} &
  \textbf{24.13} &
  \textbf{34.17} &
  \multicolumn{1}{c|}{\textbf{38.30}} &
  \textbf{17.27} &
  \textbf{21.53} &
  \multicolumn{1}{c|}{\textbf{23.05}} &
  \textbf{19.09} &
  \textbf{17.60} &
  \textbf{15.56} \\
\multicolumn{1}{l|}{SAMRank (O)} &
  33.96 &
  39.35 &
  \multicolumn{1}{c|}{39.73} &
  24.08 &
  33.40 &
  \multicolumn{1}{c|}{37.53} &
  15.28 &
  18.36 &
  \multicolumn{1}{c|}{18.03} &
  16.35 &
  15.91 &
  14.52 \\ \hline
\multicolumn{13}{c}{GPT-2 Model} \\ \hline
\multicolumn{1}{l|}{SAMRank (B)} &
  33.96 &
  38.24 &
  \multicolumn{1}{c|}{38.11} &
  23.99 &
  32.51 &
  \multicolumn{1}{c|}{35.75} &
  17.23 &
  20.76 &
  \multicolumn{1}{c|}{21.56} &
  13.37 &
  15.19 &
  14.27 \\
\multicolumn{1}{l|}{Attention-Seeker} &
  \textbf{34.14} &
  \textbf{39.46} &
  \multicolumn{1}{c|}{38.89} &
  24.30 &
  33.83 &
  \multicolumn{1}{c|}{37.16} &
  \textbf{18.82} &
  \textbf{23.08} &
  \multicolumn{1}{c|}{\textbf{23.73}} &
  \textbf{19.89} &
  \textbf{17.95} &
  \textbf{15.80} \\
\multicolumn{1}{l|}{SAMRank (O)} &
  33.92 &
  39.44 &
  \multicolumn{1}{c|}{\textbf{39.72}} &
  \textbf{24.80} &
  \textbf{34.75} &
  \multicolumn{1}{c|}{\textbf{38.78}} &
  15.88 &
  18.36 &
  \multicolumn{1}{c|}{19.03} &
  17.49 &
  16.46 &
  14.92 \\ \hline
\end{tabular}
\caption{
    Performance of keyphrase extraction of SAMRank and Attention-Seeker on Small Language Models. SAMRank (B):non-parametric base method. SAMRank (O): optimal manual selection \citep{samrank}.
  }
\label{tab:tab_apx3}
\end{table*}

Finally, Attention-Seeker outperforms the Global Score of SAMRank in long documents, highlighting the importance of our proposed relevance score for document segments $T^{s}$. While this score proves crucial for extracting keyphrases from long documents, the results also respond to the limitation of SAMRank in selecting a single SAM for all document segments. When different segments have their own optimal SAM, this constraint leads to suboptimal results. In addition, the reduced likelihood of selecting a non-relevant SAM contributes to a more stable selection with lower performances. This stability is reflected in the results of the SemEval2010 and Krapivin datasets in Table ~\ref{tab:tab_apx1}, where in 3 out of 4 cases, the selected SAM is consistently from layer 12, head 2.

\section{Performance in Small Language Models: BERT and GPT-2}
\label{seg:appx_cr1}

We replicated the evaluation of Attention-Seeker and the non-parametric SAMRank method (Base) from the main paper, applying our method to smaller Language Models, specifically BERT \citep{bert} and GPT-2 \citep{gpt2}. The results of this evaluation are summarized in Table~\ref{tab:tab_apx3}.

These results align with those obtained using the LLAMA 3-8B implementation (Table~\ref{tab:tab_2}) and indicate that Attention-Seeker always outperforms the non-parametric SAM (base model that treats all layers and heads as equally relevant; Appendix~\ref{sec:apx_nonp_sam}). These findings emphasize the importance of considering the relevance of an LLM's layers and heads in the process of keyphrase extraction.

Similar to the analysis in Appendix~\ref{sec:apx_optimal_sam}, we further compare the results of Attention-Seeker with the optimal performance of SAMRank reported by \citet{samrank}. For short documents, the optimal results from SAMRank suggest room for improvement in our method, while for long documents, Attention-Seeker consistently outperforms SAMRank. Although these results are similar to those obtained with the LLAMA 3-8B implementation, Attention-Seeker outperforms the optimal SAMRank in a significant number of cases for short documents. These results suggest that the simplicity of our hypothesis vector is more effective when using smaller models, while larger models may require more sophisticated vectors to achieve optimal performance.

\section{Results of Non-Parametric SAMRank}
\label{sec:apx_nonp_sam}

The non-parametric SAMRank introduced in this study implements the method proposed by SAMRank \citep{samrank}. However, instead of selecting a single SAM from the heads of the LLM (as in the original parameterized method), it averages all SAMs. Given the set of SAMs $A = \{A^{lh},\forall l,h\}$, the averaged SAM $A^{av}$ is calculated as shown in Equation ~\ref{eq:eq_apx1} ($p$=$q$=32 for LLAMA 3-8B).

\begin{equation}
  \label{eq:eq_apx1}
A^{av}=\frac{1}{p*q}\sum_{l=1}^{p}\sum_{h=1}^{q}A^{lh}
\end{equation}

The global score is calculated as shown in Equation ~\ref{eq:eq_apx2} ($n$ is the number of tokens of the document or segment).

\begin{equation}
  \label{eq:eq_apx2}
G=\sum_{i=1}^{n}\sum_{j=1}^{m}A^{av}_{ij}
\end{equation}

The proportional score is calculated as shown in Equation ~\ref{eq:eq_apx3}. Here, $A^{x}$ is equal to the matrix $A^{av}$, where its elements $A^{av}_{i0}$ are set to zero.

\begin{equation}
  \label{eq:eq_apx3}
P=\sum_{j=1}^{n}\frac{A^{x}_{ij}}{\sum_{i=1}^{n}A^{x}_{ij}}
\end{equation}

Equation ~\ref{eq:eq_apx3} is a simplification of the equations proposed by \citet{samrank}, as their redistribution of the global score $G$ is cancelled out by the column normalization of the matrix. The only observable effect of this redistribution is that all elements $A^{av}_{i0}$ become zero, since the global score $G$ for the first token is zero.

 \begin{figure*}[h]
  \centering
  \includegraphics[width=\linewidth]{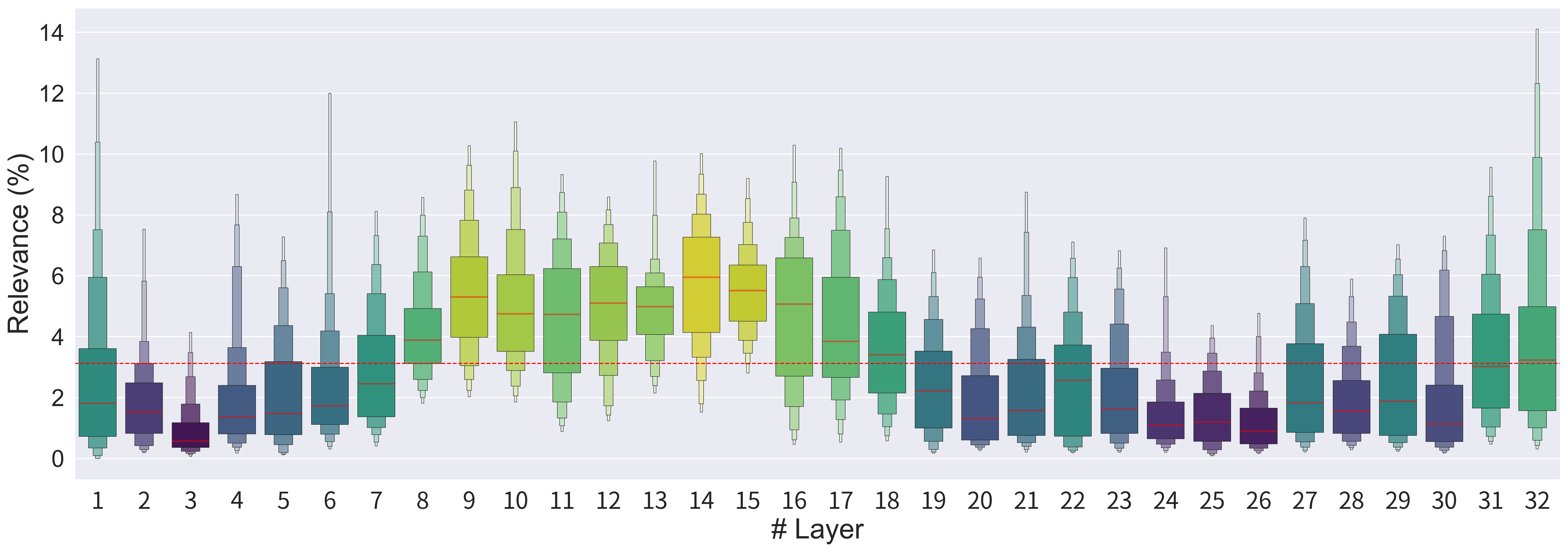}
  \caption{Relevance of LLAMA 3-8B layers for keyphrase extraction estimated by Attention-Seeker over four datasets. (The lighter the color, the higher the relevance rank of the corresponding layer)}
  \label{fig:fig_apx1}
\end{figure*}

The final score $S$ is determined by the sum of the global score $G$ and the proportional score $P$, as shown in Equation ~\ref{eq:eq_apx4}. The score for each candidate phrase is derived from the attention scores of this vector, as detailed by \citet{samrank}.

\begin{equation}
  \label{eq:eq_apx4}
S=G+P
\end{equation}

\begin{table}[h]
\fontsize{10.5pt}{13.25pt}\selectfont
\addtolength{\tabcolsep}{-2pt}
\centering
\begin{tabular}{lccc}
\hline
\multicolumn{1}{l|}{\textbf{Dataset}}     & \textbf{F1@5} & \textbf{F1@10} & \textbf{F1@15} \\ \hline
\multicolumn{4}{c}{SAMRank Global Score (G)}                                                \\ \hline
\multicolumn{1}{l|}{\textbf{Inspec}}      & 34.25         & 38.18          & 38.11          \\
\multicolumn{1}{l|}{\textbf{SemEval2017}} & 24.74         & 33.51          & 37.01          \\
\multicolumn{1}{l|}{\textbf{SemEval2010}} & 17.86         & 20.99          & 22.07          \\
\multicolumn{1}{l|}{\textbf{Krapivin}}    & 17.38         & 16.78          & 15.15          \\ \hline
\multicolumn{4}{c}{SAMRank Proportional Score (P)}                                          \\ \hline
\multicolumn{1}{l|}{\textbf{Inspec}}      & 27.21         & 33.81          & 35.20          \\
\multicolumn{1}{l|}{\textbf{SemEval2017}} & 20.49         & 25.90          & 33.01          \\
\multicolumn{1}{l|}{\textbf{SemEval2010}} & 15.75         & 19.07          & 19.34          \\
\multicolumn{1}{l|}{\textbf{Krapivin}}    & 14.84         & 14.45          & 13.27          \\ \hline
\multicolumn{4}{c}{SAMRank Final Score (S)}                                                 \\ \hline
\multicolumn{1}{l|}{\textbf{Inspec}}      & 30.23         & 36.62          & 38.08          \\
\multicolumn{1}{l|}{\textbf{SemEval2017}} & 21.95         & 31.83          & 35.39          \\
\multicolumn{1}{l|}{\textbf{SemEval2010}} & 16.36         & 20.13          & 20.87          \\
\multicolumn{1}{l|}{\textbf{Krapivin}}    & 16.06         & 15.46          & 14.29          \\ \hline
\end{tabular}
\caption{
    Performance of keyphrase extraction of the three scores of a non-parametric SAMRank on four datasets.
  }
\label{tab:tab_apx2}
\end{table}

We implemented the non-parametric SAMRank method using the model LLAMA 3-8B, evaluating their three different scores to ensure that the non-parametric version benefits from these scores similarly to the original SAMRank. Table ~\ref{tab:tab_apx2} shows the performance achieved by the non-parametric SAMRank in the three cases.

We observe a similar result than in the optimized implementation of SAMRank. The Proportional Score degrades the improvement of the overall performance, causing  the implementation with the Global Score to achieve the best results. Accordingly, the performance reported in Table ~\ref{tab:tab_2} is collected from the implementation of SAMRank that only use the Global Score.

\section{Most Relevant SAMs identified by Attention-Seeker}
\label{sec:apx_rel_sam}

Attention-Seeker estimates the relevance of SAMs, their internal attention vectors, and their corresponding segments. While analyzing the relevance of attention vectors and document segments requires a deep study of different cases (influence of both the input text and the selected SAM), the relevance of SAMs is more closely related to the model characteristics. Specifically, the layers of LLMs represent different levels of text processing, with higher layers capturing more complex token relationships. To investigate the impact of different layers on keyphrase extraction, we extracted the relevance scores $R^{lh}$ estimated by Attention-Seeker across all samples of the four datasets used in our study (Inspec, SemEval2017, SemEval2010, and Krapivin). 

Figure ~\ref{fig:fig_apx1} shows the distribution of $R^{lh}$ scores for each layer, with the scores normalized for consistent comparison. For better visualization, the top 5\% of outliers are filtered out from each distribution, distributions are color-coded according to their relevance rank, and a red line is added at $Relevance=3.125$ to indicate a uniform distribution where all layers are equally relevant. The analysis shows that the heads between layers 8 and 18, especially layers 9, 12, 14, and 15, received the highest relevance scores. This result suggests that heads located in intermediate layers of an LLM tend to specialize in processing tasks related to noun-noun relationships and selecting the most relevant nouns/phrases.

The first and last layers show high variability in relevance scores across heads, with some receiving remarkably high scores. This suggests that attention to keyphrases is a fundamental task: At low levels, it helps to generate well-contextualized embeddings, while at high levels, it contributes to building a complete representation of the document context. At both levels, however, this process is complemented by numerous other tasks performed by different heads. 

 \begin{figure}[t]
  \centering
  \includegraphics[width=\linewidth]{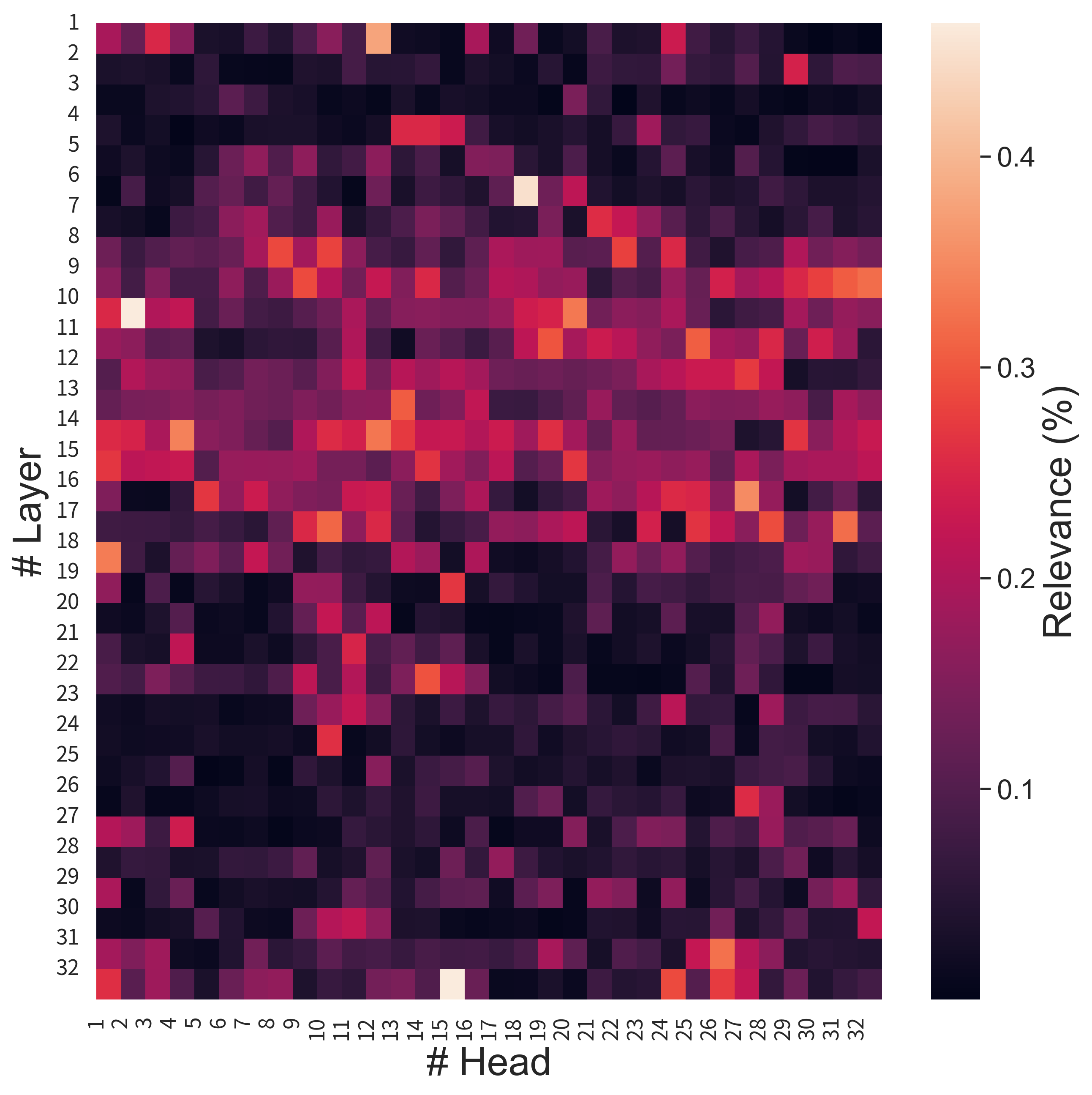}
  \caption{Relevance of LLAMA 3-8B heads for keyphrase extraction estimated by Attention-Seeker in one sample of the Inspec dataset.}
  \label{fig:fig_apx2}
\end{figure}

 \begin{figure}[h]
  \centering
  \includegraphics[width=\linewidth]{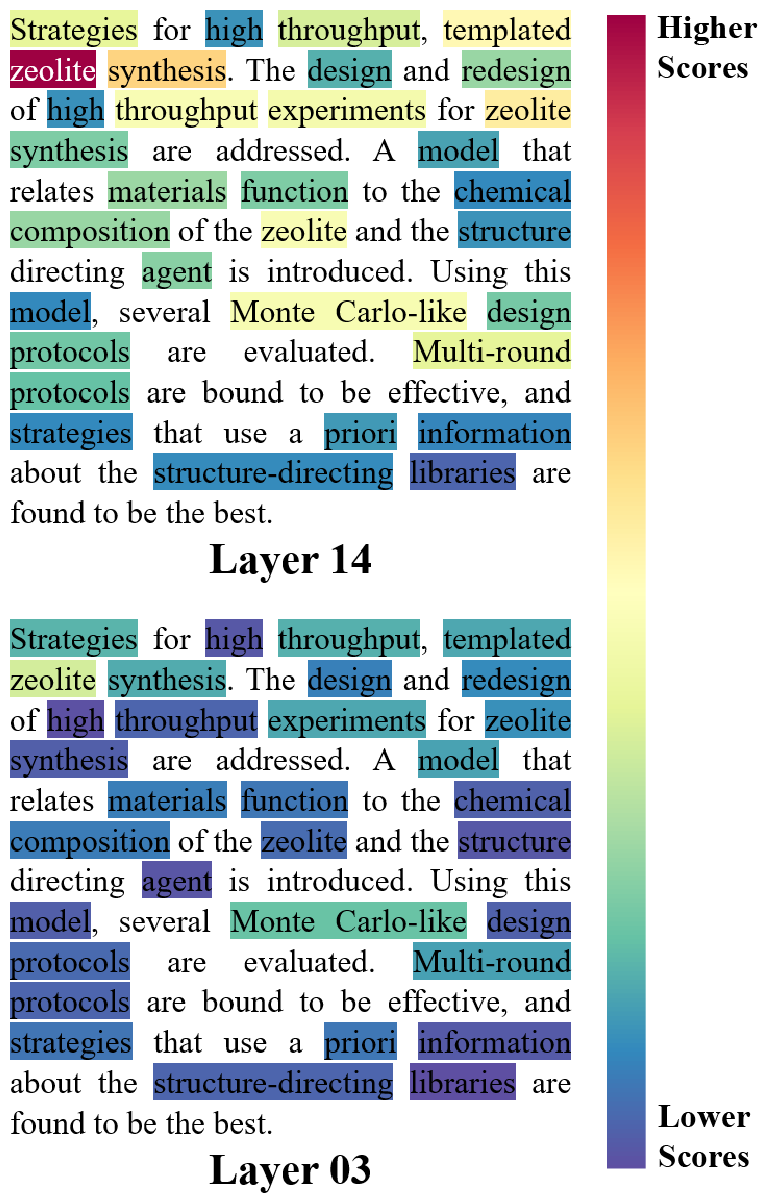}
  \caption{Attention scores in Layers 3 and 14 in one sample of the Inspec dataset.}
  \label{fig:fig_apx3}
\end{figure}

Figure ~\ref{fig:fig_apx2} supports the previous observation by illustrating the relevance scores $R^{lh}$ assigned to each head for a sample from the Inspec dataset. The score distribution across layers aligns with the general pattern in Figure ~\ref{fig:fig_apx1}, where most heads in the intermediate layers contribute significantly to keyphrase extraction, while only a few heads in the first and last layers exhibit high relevance. For the same Inspec sample, we compared the attention scores of candidate phrase tokens in layer 3 (lowest overall relevance score) and layer 14 (highest overall relevance score). Figure ~\ref{fig:fig_apx3} shows the difference between these two layers: Layer 14 assigns higher attention scores to all candidates, with a clearer contrast between phrases, while layer 3 tends to focus on non-candidate tokens, resulting in lower, more uniformly distributed attention scores.

\begin{table*}[t]
\centering
\begin{tabular}{cccccc}
\hline
\multicolumn{1}{c|}{\textbf{Model}} & \multicolumn{1}{c|}{\textbf{Method}} & \textbf{Inspec} & \textbf{SemEval2017} & \textbf{SemEval2010} & \textbf{Krapivin} \\ \hline
\multicolumn{6}{c}{Small-sized Language Models}                                                                      \\ \hline
\multicolumn{1}{c|}{\multirow{2}{*}{BERT}}      & \multicolumn{1}{c|}{SAMRank}        & 0.04 & 0.06 & 6.76  & 6.74  \\
\multicolumn{1}{c|}{}                           & \multicolumn{1}{c|}{Attention-Rank} & 0.12 & 0.13 & 7.89  & 7.89  \\ \hline
\multicolumn{1}{c|}{\multirow{2}{*}{GPT-2}}     & \multicolumn{1}{c|}{SAMRank}        & 0.05 & 0.06 & 7.16  & 7.33  \\
\multicolumn{1}{c|}{}                           & \multicolumn{1}{c|}{Attention-Rank} & 0.12 & 0.13 & 7.90  & 8.11  \\ \hline
\multicolumn{1}{c|}{T5}                         & \multicolumn{1}{c|}{PromptRank}     & 2.25 & 2.51 & 3.44  & 8.85  \\ \hline
\multicolumn{6}{c}{Medium-sized Language Models}                                                                     \\ \hline
\multicolumn{1}{c|}{\multirow{3}{*}{LLAMA 3-8B}} & \multicolumn{1}{c|}{SAMRank}        & 0.22 & 0.30 & 11.98 & 12.05 \\
\multicolumn{1}{c|}{}                           & \multicolumn{1}{c|}{Attention-Rank} & 0.68 & 0.76 & 14.00 & 14.01 \\
\multicolumn{1}{c|}{}                           & \multicolumn{1}{c|}{PromptRank}     & 5.88 & 9.89 & 45.12 & 42.82 \\ \hline
\end{tabular}
\caption{
    Average computation time (seconds) required by SAMRank, Attention-Rank, and PromptRank on small and medium-sized Language Models to process one document (batch size = 1).
  }
\label{tab:tab_apx4}
\end{table*}

\section{Computational cost}
We evaluated the computation time required for the three methods with the best general performance in unsupervised keyphrase extraction: SAMRank, PromptRank, and Attention-Seeker. The computations were performed on an NVIDIA RTX A6000 GPU with 48 GB of memory. In all implementations, documents were processed individually (batch size = 1). We conducted the evaluation using four Language Models: BERT, GPT-2, and T5 for small-sized models, and LLAMA 3-8B for medium-sized models.

The results are shown in Table~\ref{tab:tab_apx4}. PromptRank generally requires more computation time than the other two methods. This is due to its specific processing requirements: it extracts the logit for each "document" + "candidate" combination, effectively processing almost the same text $n$ times, where $n$ is the number of keyphrase candidates. A potential solution is to process all combinations as a batch, but this would shift the computational burden from time to memory resources.

Among the attention-based methods, SAMRank shows better computational efficiency compared to Attention-Seeker. This difference is expected because Attention-Seeker builds on SAMRank by introducing additional computations, including vector similarity calculations and linear transformations of SAMs. Despite these additional steps, both methods have comparable memory consumption, are efficient enough for real-time processing of short documents, and show similar computation times for long documents. Due to its superior performance, Attention-Seeker remains the preferred option overall. However, in scenarios where processing speed is critical for short documents and a small trade-off in F-score (about 2\%) is acceptable, the non-parametric SAMRank would be the better choice.



\end{document}